%% file: conference_101719.tex
\documentclass[10pt]{IEEEtran}
\IEEEoverridecommandlockouts
\usepackage{cite}
\usepackage{amsmath,amssymb,amsfonts}
\usepackage{graphicx}
\usepackage{booktabs}
\usepackage{textcomp}
\usepackage{xcolor}
\usepackage{tabularx} 
\usepackage{booktabs} 
\renewcommand{\arraystretch}{1.2} 
\usepackage[top=1in, bottom=0.8in, left=0.75in, right=0.75in]{geometry}
\usepackage[hidelinks]{hyperref}
\usepackage{caption}
\usepackage{subcaption}
\usepackage{makecell}
\usepackage{algorithm}
\usepackage{algpseudocode}
\usepackage{calculator}
\usepackage{float}
\def\BibTeX{{\rm B\kern-.05em{\sc i\kern-.025em b}\kern-.08em
    T\kern-.1667em\lower.7ex\hbox{E}\kern-.125emX}}
\begin{document}

\title{Neuro-LIFT: A Neuromorphic, \underline{L}LM-based \underline{I}nteractive \underline{F}ramework for Autonomous Drone Fligh\underline{T} at the Edge}

\author{Amogh Joshi\IEEEauthorrefmark{1}\thanks{\IEEEauthorrefmark{1}These authors contributed equally to this work.}, Sourav Sanyal\IEEEauthorrefmark{1}  and Kaushik Roy \\
  \IEEEauthorblockA{Electrical and Computer Engineering,
 Purdue University\\
 \{joshi157, sanyals, kaushik\}@purdue.edu
 }
}
\maketitle

\begin{abstract}
The integration of human-intuitive interactions into autonomous systems has been limited. Traditional Natural Language Processing (NLP) systems struggle with context and intent understanding, severely restricting human-robot interaction. Recent advancements in Large Language Models (LLMs) have transformed this dynamic, allowing for intuitive and high-level communication through speech and text, and bridging the gap between human commands and robotic actions.
Additionally, autonomous navigation has emerged as a central focus in robotics research, with artificial intelligence (AI) increasingly being leveraged to enhance these systems. However, existing AI-based navigation algorithms face significant challenges in latency-critical tasks where rapid decision-making is critical. Traditional frame-based vision systems, while effective for high-level decision-making, suffer from high energy consumption and latency, limiting their applicability in real-time scenarios. Neuromorphic vision systems, combining event-based cameras and spiking neural networks (SNNs), offer a promising alternative by enabling energy-efficient, low-latency navigation. Despite their potential, real-world implementations of these systems, particularly on physical platforms such as drones, remain scarce.
In this work, we present Neuro-LIFT, a real-time neuromorphic navigation framework implemented on a Parrot Bebop2 quadrotor. Leveraging an LLM for natural language processing, Neuro-LIFT translates human speech into high-level planning commands which are then autonomously executed using event-based neuromorphic vision and physics-driven planning. Our framework demonstrates its capabilities in navigating in a dynamic environment, avoiding obstacles, and adapting to human instructions in real-time. Demonstration images of Neuro-LIFT navigating through a moving ring in an indoor setting is provided, showcasing the system’s interactive, collaborative potential in autonomous robotics.

\begin{IEEEkeywords}
Large Language Models, Edge Intelligence, Neuromorphic Vision, Dynamic Vision Sensors, Autonomous Navigation, Physics-based AI
\end{IEEEkeywords}


\end{abstract}


\section{Introduction}
\input{sections/introduction}

\section{Preliminaries}

\input{sections/background}

\section{System Overview }
\input{sections/method}
\section{Experimental Results}
\input{sections/result}

\section{Related Work}
\input{sections/lit_review}

\section{Conclusion}
\input{sections/conclusion}

\section{Acknowledgement}

This work was supported by IARPA, and by the Center for the Co-Design of Cognitive Systems (CoCoSys), a center in JUMP 2.0, an SRC program sponsored by DARPA.

\bibliographystyle{IEEEtran}
\bibliography{conference_101719}


\end{document}

%% file: sections/introduction.tex
Existing autonomous navigation algorithms often operate in black box fashion, leaving the end user with little to no influence over the algorithm's decisions. Even if the inner workings of the algorithm were to be exposed to a human, the human-comprehension gap, i.e., our inability to comprehend the abstract representation spaces that these algorithms operate in precludes effective, high-level human interaction. This lack of effective interaction prevents such algorithms from developing human-esque intuitions and situational awareness, in turn preventing the learning of essential behavior such as shifting priorities in case of emergencies. This, in turn, is a major obstacle in the development of collaborative autonomous robotic systems. External methods for developing intuition, such as \cite{joshi2024shireenhancingsampleefficiency} have been attempted. However, these methods can not be used in an online fashion, limiting their use in a collaborative robotics setting.

Earlier Natural Language Processing (NLP) systems were able to provide some degree of speech and text handling capabilities. However, their effectiveness in human-robot interaction was limited by their inability to perform higher-level reasoning tasks such as context and intent understanding. This in turn, restricted NLP-based interaction to simplistic tasks such as keyword recognition, making interaction rigid and error-prone. In recent years, the widespread adoption of Large Language Models (LLMs) such as   GPT-3 \cite{NEURIPS2020_1457c0d6}, GPT-4 \cite{achiam2023gpt}, CLIP \cite{radford2021learning}, DALL-E \cite{ramesh2021zero} and PALM-E \cite{driess2023palm} has granted humans the ability to interact with complex robotic systems in intuitive ways such as speech and text. LLMs' ability to perform complex contextual and muti-modal understanding has enabled high-level interactions such as the integration of human speech with vision inputs and their conversion into machine commands.

Autonomous navigation algorithms have long been a focal point of interest in robotics research, with a recent emphasis on enhancing these systems using artificial intelligence (AI). However, existing AI-based navigation algorithms often struggle with reactive tasks like close-in obstacle avoidance, where rapid decisions are crucial. This limitation arises from the inability of traditional sensing modalities to provide necessary data quickly. Vision-based algorithms, while useful for high-level decision-making through techniques like object detection, are hindered by the high energy consumption and latency of conventional frame-based cameras, which can negate their obvious benefits.

\begin{figure*}[ht]
    \centering
    \includegraphics[width=0.8\textwidth]{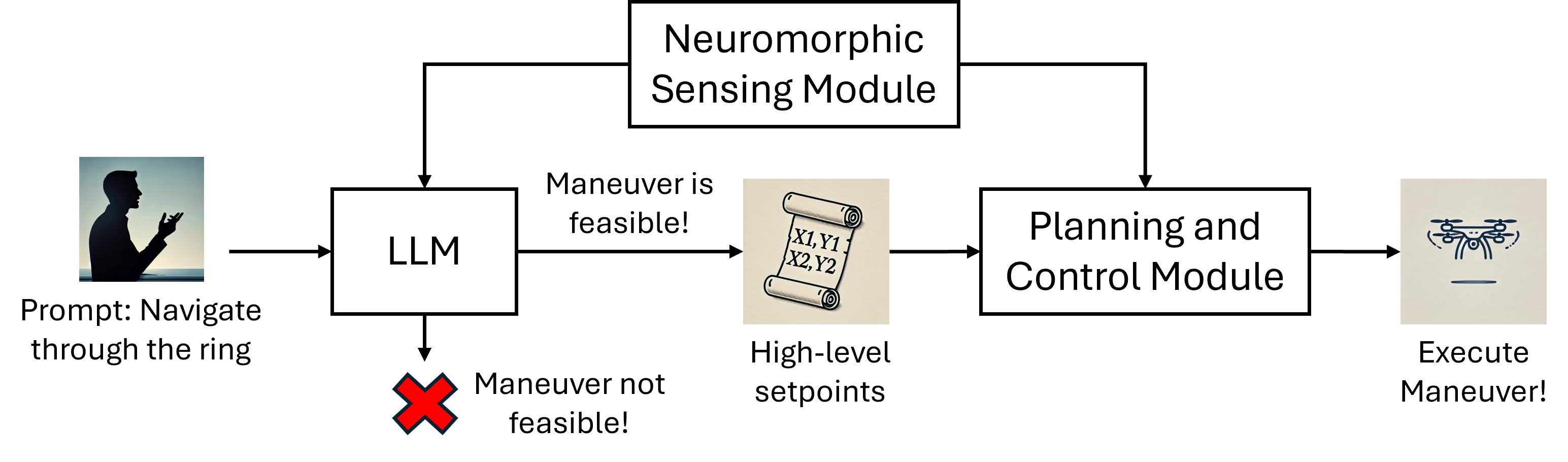}
    \caption{Functional Overview of the Neuro-LIFT Framework}
    \label{fig:funcover}
\end{figure*}

\textit{Event-based cameras}\cite{dvs1,dvs2,dvs3} offer advantages over traditional \textit{frame-based cameras} in terms of latency and energy efficiency. They only record changes in intensity that exceed a programmable threshold. This capability significantly reduces bandwidth and allows for data capture at microsecond granularity. When these cameras are paired with \textit{spiking neurons} \cite{lif}, which process information asynchronously, they become highly effective for low-energy, latency-critical applications \cite{lee2020spike, kosta2023adaptive}. Together, they support the development of \textit{neuromorphic vision} and form the technological foundation of \textit{neuromorphic navigation algorithms}, facilitating energy-efficient and responsive autonomous navigation solutions \cite{sanyal2024ev}. Despite their potential, real-world implementations of these systems are quite rare, their deployment on physical robots like drones and integrating them with physics-driven planning methods for complex operations even more so. 

In this work, we introduce Neuro-LIFT --  a neuromorphic, LLM-based interactive  navigation framework on a Parrot Bebop2 quadrotor. Neuro-LIFT utilizes an LLM (a LLaMa 3 model \cite{grattafiori2024llama3herdmodels}) to convert human-specified natural language instructions into a high-level drone-control task, and then use event-based neuromorphic vision and physics-driven planning to autonomously execute the task. To the best of our knowledge, this is the first work in reported literature that integrates LLMs with the neuromorphic modality to achieve interactive, and autonomous indoor flight. A high-level overview of our framework is as shown in Fig. \ref{fig:flowchart}. In the example shown in Fig \ref{fig:funcover}, we ask Neuro-LIFT, our LLM drone pilot to navigate a Parrot Bebop2 through/around a moving ring in an indoor environment while avoiding collisions. 


%% file: sections/background.tex
\subsection{\textbf{Human-Robot Interaction}}
Human-Robot Interaction (HRI) is the study of interactions between humans and robots. It encompasses the design, development, and evaluation of robotic systems that can work with, assist, or coexist with humans in various settings. In the context of autonomous robotics, the primary objective of HRI is to design autonomous robots that can interpret human intentions, adapt to dynamic environments, and respond effectively to real-time instructions. This requires developing intuitive communication interfaces such as natural language commands or gestures, that allow humans to guide and interact with robots effortlessly. Previous approaches to such interaction relied mainly on Natural Language Processing (NLP) techniques. However, the rigidity of conventional NLP techniques made this approach rigid and error prone. The integration of Large Language Models (LLMs) provides machines with a fast, easy, and flexible way of understanding real-time human commands and developing situational awareness, thereby enabling humans to guide autonomous robots more effectively. In the case of autonomous flight specifically, this can help bridge the human comprehension gap, allowing for the autonomous execution of a human-specified maneuver.

\subsection{\textbf{Neuromorphic Vision}}

Neuromorphic vision relies on a new type of imaging sensor called a Dynamic Vision Sensor (DVS) \cite{dvs1, dvs2, dvs3} to feed a specialized asynchronous neural network called a Spiking Neural Network (SNN). A DVS, also called an event camera, records an ``event" when the logarithmic intensity of light incident on a pixel changes by more than a preset threshold. Events are recorded at the pixel granularity, in the form of $(x, y, t, p)$ tuples, where $x,y$ are the co-ordinates (in pixel space) of the event, $t$ is the event's timestamp, and $p$ is the polarity of the event, i.e., it denotes whether the light intensity increased or decreased. This fully asynchronous data recording format results in very high temporal resolution, while still keeping bandwidth requirements low as unlike frame-based cameras, events are not generated at every pixel simultaneously. Thus, an event camera can be described as a very low latency, high temporal resolution image sensor. Event cameras offer several distinctive advantages due to their unique operating principles. One significant benefit is their high dynamic range, which is achieved because events are generated solely due to logarithmic changes in intensity. This characteristic makes event cameras extremely robust to sudden variations in lighting conditions. Another advantage is their low redundancy; events are triggered only by changes in intensity. Consequently, pixels that do not experience any change in intensity do not generate events, which greatly reduces the amount of redundant information captured.

Another key aspect of neuromorphic vision is its use of \textit{spiking neurons} arranged spatially to form a Spiking Neural Network (SNN). Spiking neurons are a class of specialised bio-plausible neurons that accumulate incoming spikes into their ``membrane potential", producing an output spike only when the membrane potential reaches a ``threshold" value. Additionally, these neurons possess a ``leak" factor that reduces (``leaks") part of the membrane potential at every timestep. Thus, SNNs have an inherent ability to filter out noise, as membrane potential generated by noise inevitably leaks out before it can be topped up by additional noise.
Recent works such as DOTIE \cite{nagaraj2023dotie} and TOFFE \cite{toffe} have shown that carefully designed configurations of SNNs can significantly outperform conventional architectures at the object detection task, which is one of the most important tasks for autonomous navigation.

\subsection{\textbf{Physics-Based AI}}
Physics-aware AI refers to artificial intelligence systems designed to incorporate the principles of physics such as dynamics, kinematics, energy conservation, and environmental constraints into the learning process in order to enhance the decision-making capabilities of an autonomous AI agent. Unlike traditional AI models that rely purely on data-driven approaches, physics-aware AI combines these models with an understanding of the physical world, allowing robots to reason about and adapt to their environments and predict the outcome of their actions more effectively, thereby enabling the safe, efficient, and reliable operation of robots in dynamic and often unstructured environments.


\subsection{\textbf{Our Approach - Best of all Worlds}}
Modular frameworks for autonomous, interactive flight have the advantage of being easy to explain and maintain without sacrificing performance or efficiency. Therefore, in this work, we take a best-of-all-worlds approach and combine the three modules mentioned previously into a unified, modular framework for intuitively explainable, interactive, and autonomous indoor flight.

%% file: sections/method.tex
In this work, we deploy a fine-tuned LLM customized for our drone along with a modified real-time adaptation of the neuromorphic navigation algorithm EV-Planner \cite{sanyal2024ev} on a quadrotor drone using event-based vision and physics-driven planning. To highlight the robustness of our implementation, we randomize the velocity of the dynamic obstacle and also add randomized delays at various points in our framework. Section \ref{funcover} provides a functional overview of our setup. The components of our setup are as shown in Table \ref{tab:components}, and are explained in greater detail in sections \ref{sensemod}-\ref{plancont}.

\begin{table*}[!t]
    \centering
        \caption{Neuro-LIFT: Individual Component Specifications}
    \begin{tabular}{c c c}
    
    \toprule
        Module Name & Components & Specifications \\
        \midrule
        \textbf{Human Interaction} & LLM & \makecell{LLaMA-3.2-3B Instruct LLM model \cite{grattafiori2024llama3herdmodels} fine-tuned in few-shot fashion\\on custom dataset as described in Section \ref{humint}} \\
        \textbf{Neuromorphic Sensing} & \makecell{DVS Sensor\\Depth Sensor\\IMU} & \makecell{Davis346B Sensor, Spatial Resolution: 346x260 pix, Temporal Resolution: $20\mu s$\\Intel Realsense Depth Camera, Frame Rate: 30fps, Max Depth: 10m\\MPU9250 9 DoF IMU} \\
        \textbf{Planning and Control} & \makecell{Planner\\Low-level Control} & \makecell{EV-Planner: Physics-aware Neuromorphic Planner\\Conventional PID} \\
        
        \bottomrule
    \end{tabular}
    \label{tab:components}
\end{table*}

\label{funcover}
Fig. \ref{fig:funcover} diagrammatically illustrates the overall organization of our framework. Initially, a human subject asks Neuro-LIFT to perform a maneuver in an indoor environment. The fine-tuned Neuro-LIFT LLM determines the safety and feasibility of the maneuver, and, if found to be feasible, generates a high-level machine command for Ev-Planner. At this point, the drone and the teleoperated TurtleBot are armed and the TurtleBot begins moving. Simultaneously, the drone takes off and stabilizes, and upon stabilization, starts tracking the ring. Ev-Planner then generates a flight plan, which is executed using low-level Proportional-Integral-Derivative (PID) controllers. We call this real-world implementation \textit{EV-PID}. Using our engineered setup, the drone autonomously navigates through (or around) a moving ring in an indoor environment, without collision. The individual modules are explained in greater detail in the following sections.

\subsection{\textbf{Neuromorphic Sensing Module}}\label{sensemod}
\begin{figure}[ht]
     \centering
     \includegraphics[scale=0.45]{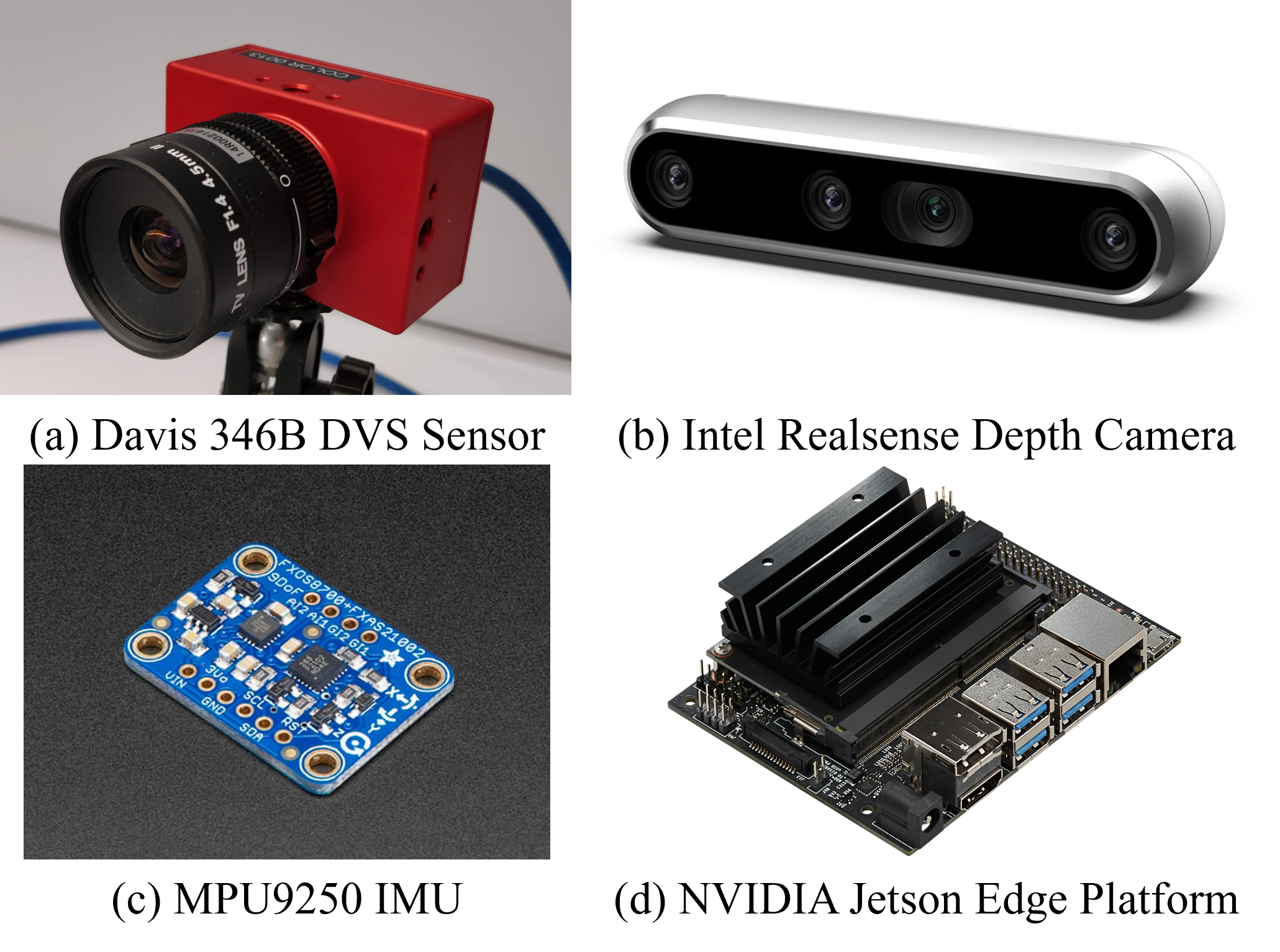}
        \caption{Components of the Neuromorphic Sensing Module}
        \label{figure:sensemodule}
\end{figure}

This module provides Neuro-LIFT and the Planning and Control module with the necessary sensory data. This module consists of a neuromorphic Event camera (Davis 346B, Fig. \ref{figure:sensemodule}a), a Depth Camera (Intel Realsense , Fig. \ref{figure:sensemodule}b), and an IMU (Fig. \ref{figure:sensemodule}c). Raw data from these sensors is processed by an NVIDIA Jetson Nano (Fig. \ref{figure:sensemodule}d) edge processor to generate the environment state report required by the LLM and feedback data as required by the planning and control module. 

\begin{figure*}[ht]
    \centering
   \includegraphics[width = 0.7\textwidth]{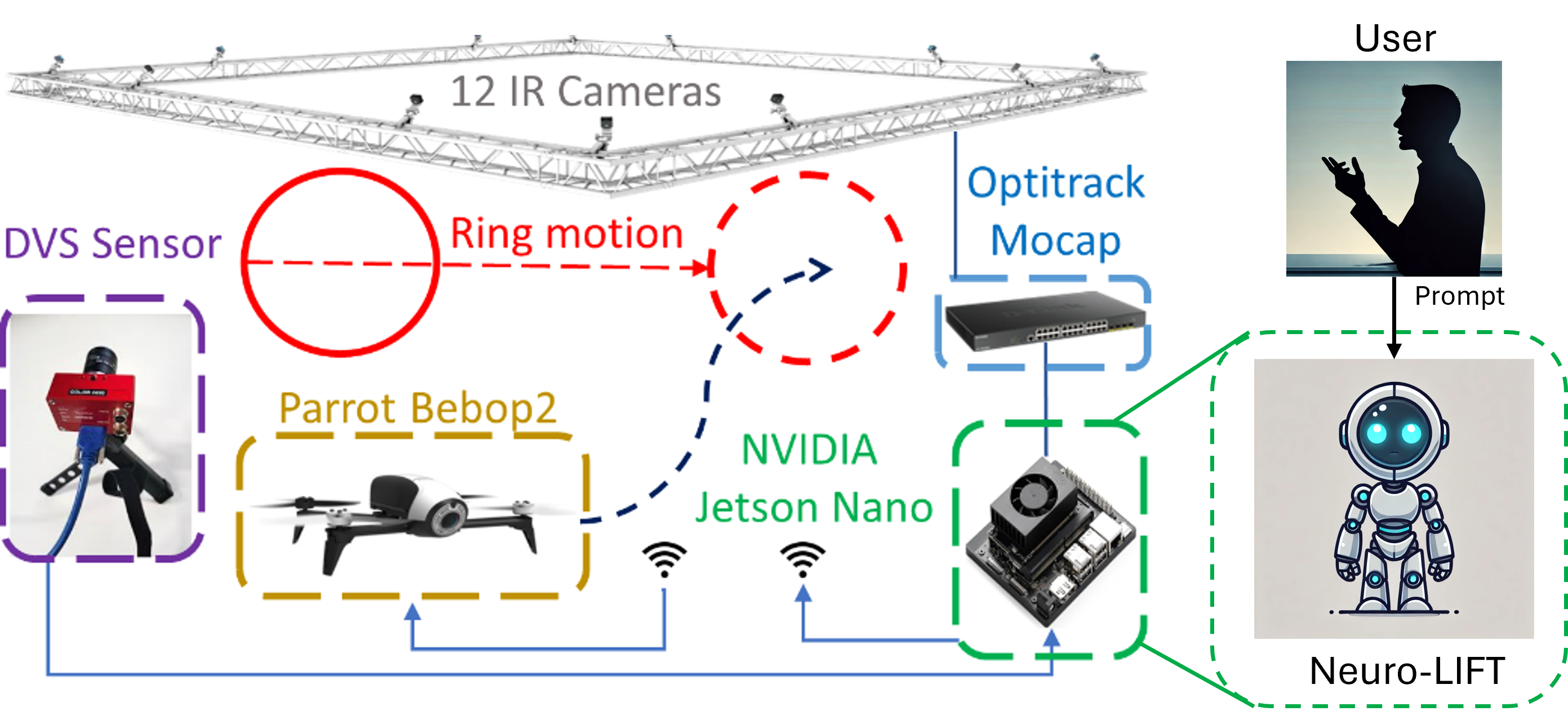} 
   \caption{Planning and control module:
    Drone and DVS sensor poses are taken from the Optitrack motion capture system consisting of 12 IR cameras. The ring is tracked using the DVS Sensor. Planning and control algorithms are executed on an Off-Board NVIDIA Jetson Edge processor, which sends control commands to the Parrot Bebop2 over a private WiFi network.}
   \label{figure:method}
\end{figure*}

\subsection{\textbf{Human Interaction Module}}\label{humint}
The human interaction component of Neuro-LIFT is a fine-tuned Large Language Model (LLM) (Llama-3.2-3B Instruct) \cite{grattafiori2024llama3herdmodels} developed to enhance intuitive command and control capabilities within our neuromorphic drone framework. Unlike Vision-Language Models (VLMs), which require large datasets and significantly more computational resources to fine-tune, Neuro-LIFT focuses on leveraging the relative computational efficiency of fine-tuning LLMs. Our approach not only simplifies the architecture of the system by removing computationally intensive vision models and other redundant components, but also improves the ability of the model to generate precise flight plans based on complex mathematical calculations and real-time data analysis.

Our fine-tuning strategy enables Neuro-LIFT to operate with high mathematical rigor, ensuring detailed and accurate flight path planning and execution. This component uses sensor data along with constraints derived from physics, environmental knowledge, and safety bounds to determine maneuver feasibility. If a maneuver is deemed to be feasible, then Neuro-LIFT toggles a go/no-go signal to high, telling the planning and control framework to commence maneuver execution. Neuro-LIFT continuously monitors the state of the environment and the objective. If, at any point during maneuver execution, Neuro-LIFT deems further execution to be too risky, it flips its output to no-go, forcing the drone to land immediately. The Neuro-LIFT Llama model is fine-tuned to solve the binary classification problem in few-shot fashion using a custom dataset that we generated.

\noindent{\textbf{Finetuning Dataset}}: The fine-tuning dataset consists of $5000$ samples of the form $(p, esr, c)$, where $p$ is a prompt provided by the user, $esr$ is the environment state report generated by the neuromorphic sensing module, and $c$ is a list containing the drone's physical capabilities such as max safe velocity and acceleration; while the labels are simply $1$ or $0$, corresponding to the go and no-go signals. These labels are generated by checking the kinematic, velocity, and time feasibility of the required movement, as shown in Algorithm \ref{alg:feasibility}. Note that $src$ denotes quantities related to the drone, while $dest$ denotes quantities related to the setpoint to be achieved. This point corresponds to the center of the moving ring in our example.
\begin{algorithm*}[ht]
    \begin{algorithmic}
        \caption{Maneuver Feasibility Check}\label{alg:feasibility}
        \Require $p_{src}, p_{dest}, v_0, v_{max}, a_{max}, t$ \Comment{Co-ordinates of source and destination point, initial and\\\hspace{90mm}max. safe velocity, max safe acceleration, time to collision}
        \State $d \gets \sqrt{(x_{dest}-x_{src})^2+(y_{dest}-y_{src})^2+(z_{dest}-z_{src})^2}$ \Comment{distance between source and destination}
        \State $a \gets \frac{2(d-v_0t)}{t^2}$ \Comment{Kinematic feasibility}
        \If{$a > a_{max}$}
            \State \Return  False
        \EndIf
        \State $v_f \gets v_0 + at$ \Comment{Velocity feasibility}
        \If{$v_f > v_{max}$}
            \State \Return  False
        \EndIf
        \If{$d > v_{max}t$} \Comment{Time feasibility}
            \State \Return  False
        \EndIf
        \State \Return True
    \end{algorithmic}
\end{algorithm*}
To prevent the LLM from fixating on a particular prompt for each maneuver (of the form provided in the dataset), we create at least five different prompts for each maneuver. $src$ and $dest$ co-ordinates are generated in random fashion, with bounds corresponding to the physical bounds of our indoor space, while max safe velocity and acceleration values are obtained experimentally such that the roll and pitch angles required to achieve them are within ``safe" limits.
Simliar to the fine-tuning dataset, we generate a ``test dataset" containing $1000$ samples subject to the same constraints as the finetuning dataset. This dataset is used exclusively for testing the fine-tuned LLM.

\noindent{\textbf{Fine-tuning}}: To fine-tune the Neuro-LIFT Llama model, the inputs and labels from the custom dataset are tokenized, a classification head is added to the base Llama model (Llama-3.2-3B Instruct) to adapt it for sequence classification, and the model is trained for 5 epochs, yielding $97.5\%$ maneuvering accuracy (39/40 maneuvers successful). We chose to train for only 5 epochs as we observed that more epochs led to significant overfitting to our prompts. This overfitting leads to false positives on test and real world data, i.e., infeasible maneuvers are mis-classified as feasible resulting in poor real-world performance. 

\begin{figure*}[!t]
    \centering
    \includegraphics[width=0.7\textwidth]{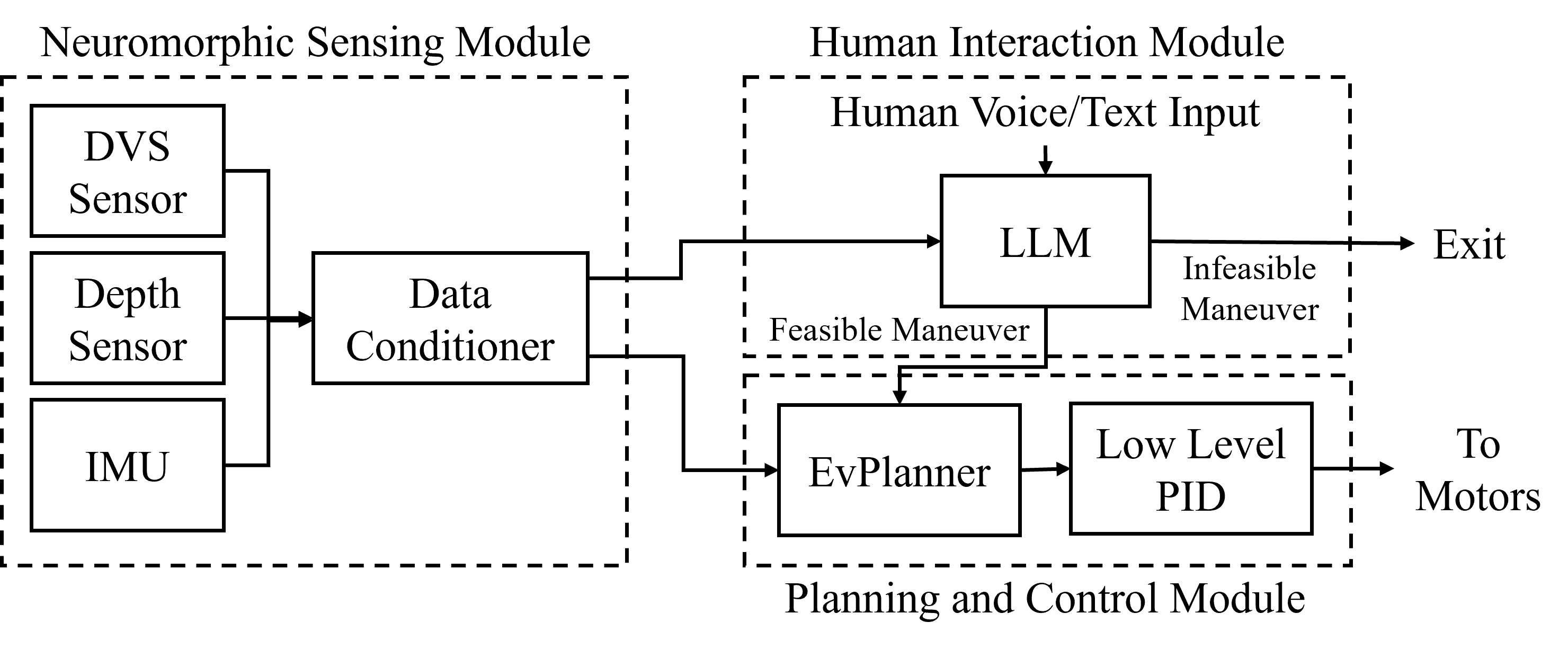}
    \caption{Neuro-LIFT Edge-AI System Architecture. EV-Planner is adapted from \cite{sanyal2024ev}.}
    \label{fig:flowchart}
\end{figure*}

\subsection{\textbf{Planning and Control Module}}\label{plancont}
As illustrated in Figure \ref{figure:method}, the planning and control module is a core component of the Neuro-LIFT system. This module utilizes a real-time implementation of the neuromorphic navigation algorithm, EV-Planner \cite{sanyal2024ev}, onboard a Parrot Bebop2 drone. The drone and its associated DVS sensor's positions are continuously updated using data from the Optitrack motion capture system, which consists of 12 IR cameras.
The EV-Planner receives high-level commands from Neuro-LIFT, translating these into detailed flight paths that take into account the dynamic motion of obstacles as captured by the DVS sensor. This setup ensures that the generated flight plans are both safe and feasible, adapting in real time to the changing environment. The flight commands are executed via low-level Proportional-Integral-Derivative (PID) controllers on the drone's flight computer, referred to as EV-PID, which communicates with the drone's motors over a private WiFi connection established by an off-board NVIDIA Jetson Nano Edge processor.
The integration of neuromorphic sensors, such as the Davis346B event camera, significantly enhances the agility and responsiveness of the navigation system while reducing overall power consumption. This optimized setup demonstrates the practical application of neuromorphic computing in enhancing the functionality and efficiency of autonomous drones.

\subsection{\textbf{System Architecture}}

The architecture of the Neuro-LIFT system is detailed in Figure \ref{fig:flowchart}. Initially, human commands are input into Neuro-LIFT through voice or text via the Human Interaction Module. Simultaneously, the Neuromorphic Sensing Module, which includes a DVS Sensor, Depth Sensor, and IMU, processes environmental data such as the count, position, and estimated velocity of obstacles. This data is conditioned and summarized before being passed to Neuro-LIFT.

Upon processing the environmental information, Neuro-LIFT assesses whether the commanded maneuver is feasible. If the maneuver is determined to be infeasible, Neuro-LIFT will notify the user and terminate the process, ensuring safety by preventing potentially hazardous actions. If feasible, Neuro-LIFT then generates a high-level flight plan that includes start and end points, along with key waypoints or states that are crucial for the successful execution of the maneuver. This flight plan is forwarded to the Planning and Control Module, which executes the maneuver using the Ev-Planner and low-level PID control that commands the drone's motors.
This structured approach allows Neuro-LIFT to integrate seamlessly with both neuromorphic hardware and sophisticated AI algorithms, optimizing the system's response time and energy efficiency while ensuring  operational safety.

\begin{figure}[!t]
    \centering
    \includegraphics[width=0.9\linewidth]{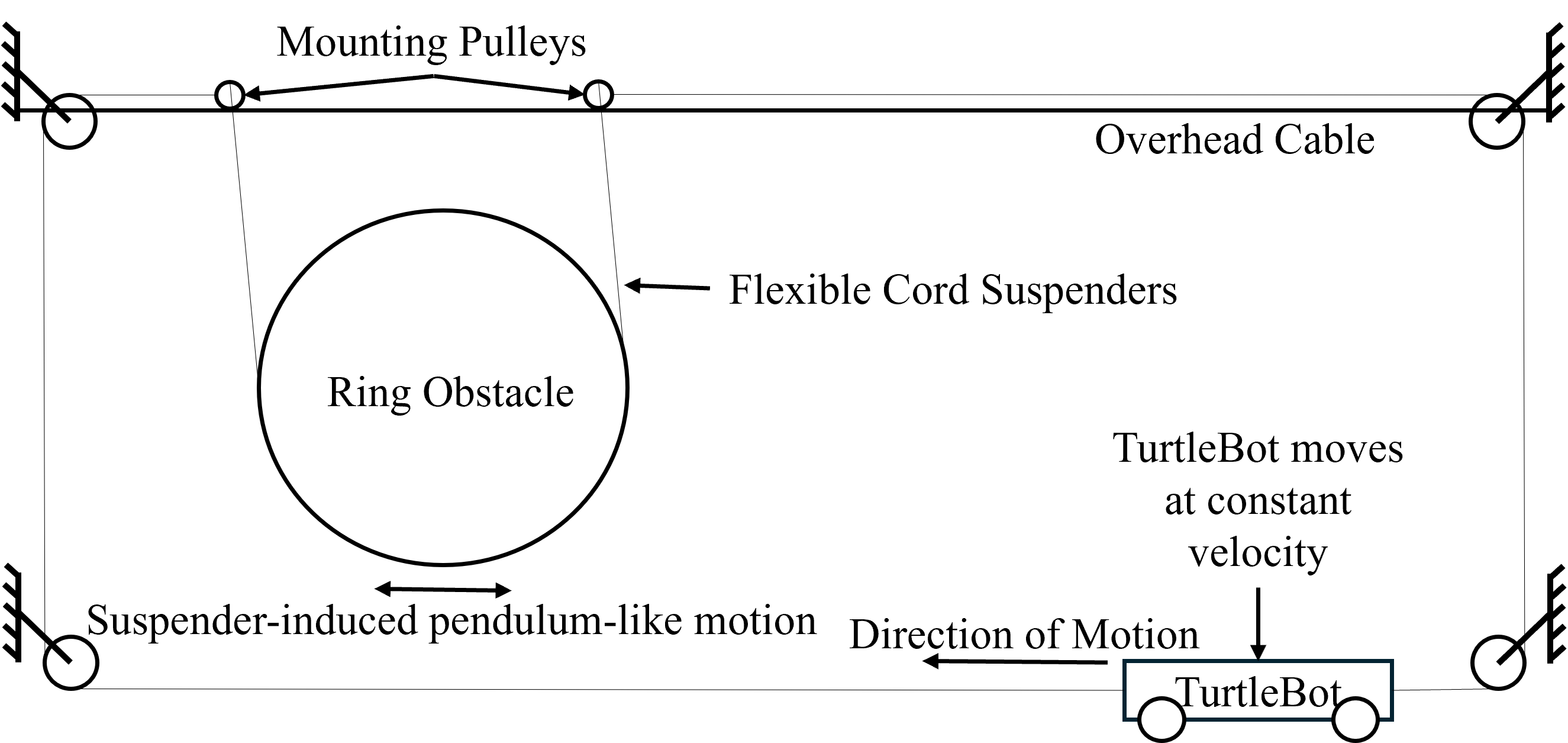}
    \caption{Pulley-and-chain mounting mechanism of the ring obstacle in our indoor environment}
    \label{figure:ringmount}
\end{figure}

\subsection{\textbf{Indoor Setup and Practical Considerations}}
Due to space constraints and safety considerations, we selected the Parrot Bebop2 quadrotor as our platform. This design choice introduced a number of challenges. The Bebop has a Maximum Takeoff Weight (MTOW) of $\sim600$gm and a deadweight of $\sim550$gm. The resulting $50$gm payload is insufficient to carry the weight of the DVS sensor (weight: $140$gm) and the edge compute device (Jetson Nano, weight: $250$gm with heatsink). Therefore, both the sensing and compute parts of the planning and control module are mounted on a tripod separate from the drone itself. The consequent relative motion between the sensor and the drone is cancelled out using drone pose feedback from a motion capture system \cite{optitrack}. Larger drones, which could carry both the DVS sensor and Jetson, do not require such compensation but are unsuitable for safe flight in our confined indoor space. Fig. \ref{figure:method} illustrates the data flow in our implementation during maneuver execution. Note that the pose feedback data is NOT used by the planner. To further prevent any unintentional feedback from the motion capture system, the velocity of the moving ring is randomized, and a random starting delay is added to the drone's flight plan.
\begin{figure}[!t]
    \centering
    \includegraphics[width=0.9\linewidth]{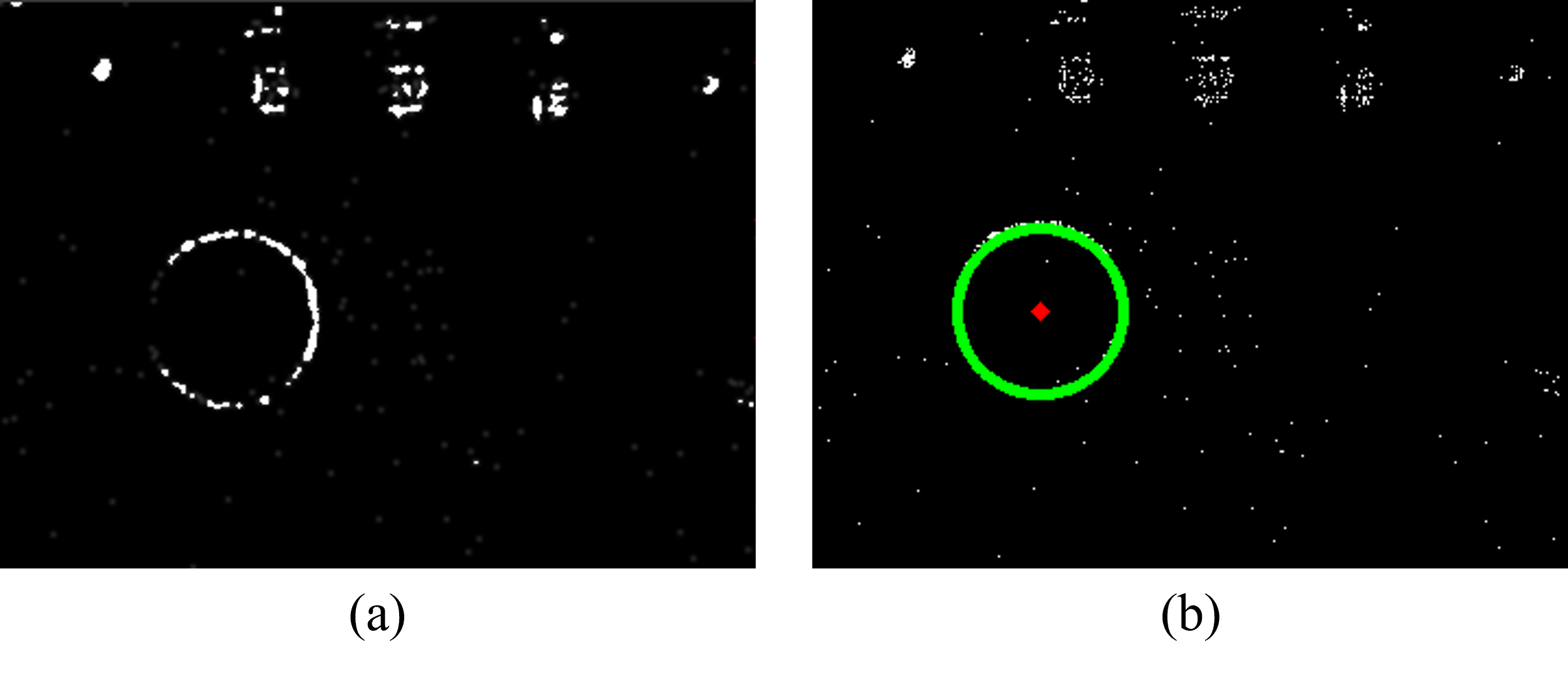} 
    \caption{Processing of event data captured by the Neuromorphic DVS Sensor. (a) Raw event data from the Neuromorphic DVS Sensor, showing the dynamic capture of environmental changes. (b) The ring detected using a shallow Spiking Neural Network, demonstrating the network's ability to interpret and visualize the significant features from the raw data.}
    \label{fig:dvs_output}
\end{figure}
The dimensions of our indoor setup necessitate driving the ring obstacle at near-constant velocity. Such a constant drive is achieved using a teleoperated TurtleBot coupled to a pulley-and-chain mechanism, as shown in Fig. \ref{figure:ringmount}. This setup enables bi-directional motion of the ring at speeds $5$cm/s - $50$cm/s. The ring obstacle is suspended from pulleys riding on an overhead cable using two other cables. This creates a pivot joint at the cable's mount point, enabling the ring obstacle to sway about the pivot. This introduces a pendulum-like movement during ring motion, ensuring that the ring has a variable velocity in spite of the constant driver velocity. These velocity variations are designed to prevent the planner from shutting down after creating a single plan at start time, while also highlighting its dynamism in the face of changing obstacle motion. 

%% file: sections/result.tex
\subsection{\textbf{Neuromorphic Object Tracking}}

Figure \ref{fig:dvs_output} showcases the capabilities of our neuromorphic vision system in detecting and tracking dynamic objects in real time. Figure \ref{fig:dvs_output} (a) displays the raw event data from the Neuromorphic DVS Sensor, which captures pixel-level changes in the scene as they occur, offering a granular view of motion dynamics. Figure \ref{fig:dvs_output}  (b) illustrates the effective utilization of a shallow Spiking Neural Network (SNN) to process these raw data inputs and accurately detect the ring structure within the scene. This processing step highlights the SNN's ability to distill relevant spatial and temporal information from the high-resolution event stream, enabling precise object localization and tracking in complex dynamic environments. The efficiency of this approach is critical for applications requiring real-time responsiveness and low power consumption, such as autonomous drone navigation and interactive robotic systems.

\begin{table*}[!t]
    \centering
    \caption{LLM Instructions for Neuromorphic Drone Navigation}
    \label{tab:llm_instructions}
    \renewcommand{\arraystretch}{1.1} 
    \setlength{\tabcolsep}{6pt} 
    \begin{tabularx}{\linewidth}{lX}
        \toprule
        \textbf{User Command} & \textbf{Neuro-LIFT Response} \\
        \midrule
        Fly Left of Ring & \textbf{Success:} "Successfully navigated around the ring from the left." \quad \textbf{Reject:} "Exiting. Preparing to land." \\
        Fly through Center of Ring & \textbf{Success:} "Successfully passed through the center of the ring." \quad \textbf{Reject:} "Exiting. Preparing to land." \\
        Fly Right of Ring & \textbf{Success:} "Successfully navigated around the ring from the right." \quad \textbf{Reject:} "Exiting. Preparing to land." \\
        \bottomrule
    \end{tabularx}
\end{table*}
\vspace{1mm}
\begin{figure*}[!t] 
    \centering
    \includegraphics[width=0.85\textwidth]{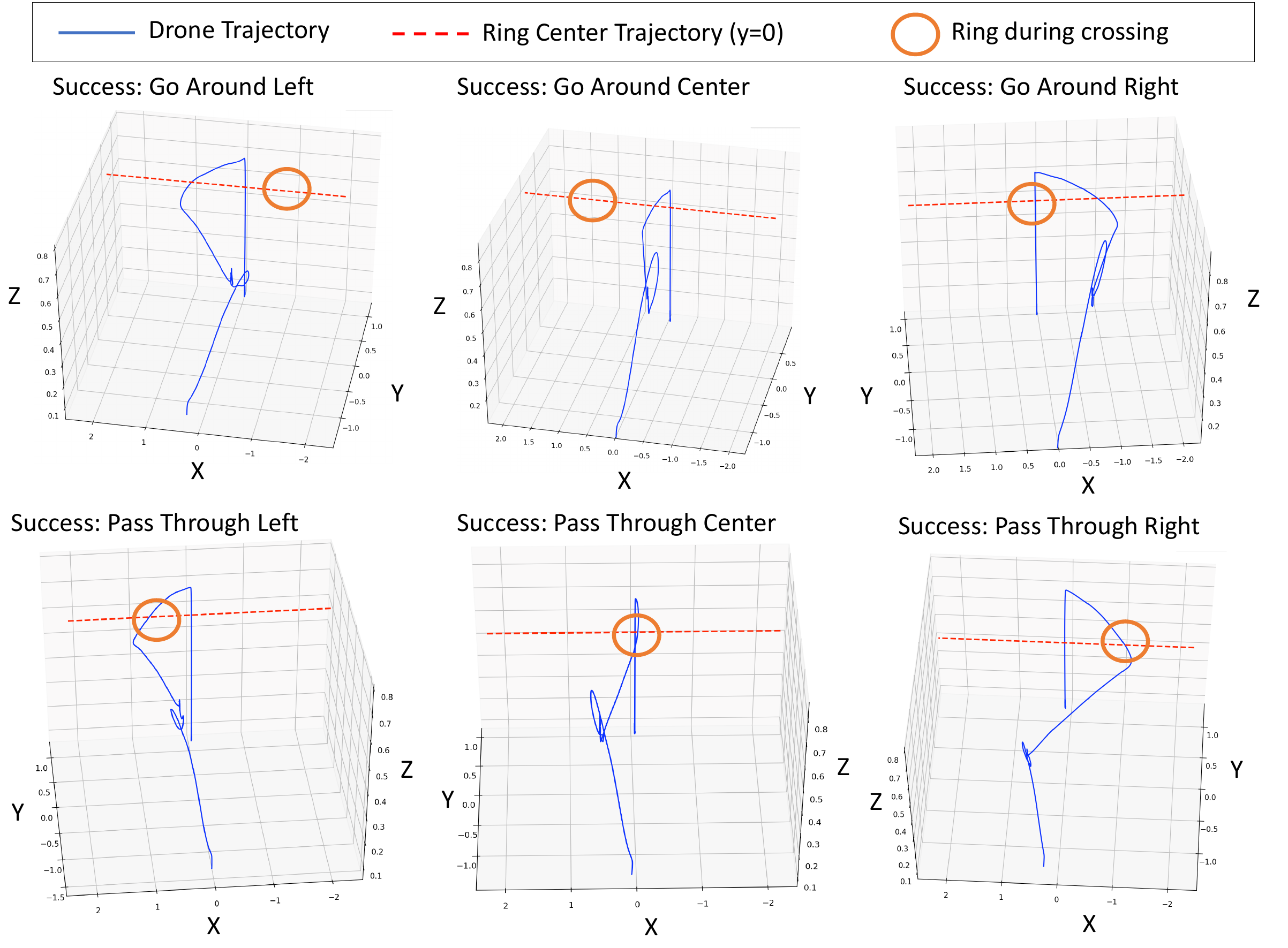} 
    \caption{Drone navigation trajectories through the ring. The success and reject cases are visualized with the ring's trajectory and the corresponding crossing points.}
    \label{fig:drone_trajectories}
\end{figure*}
\begin{figure}[!t] 
    \centering
    \includegraphics[width=0.8\linewidth]{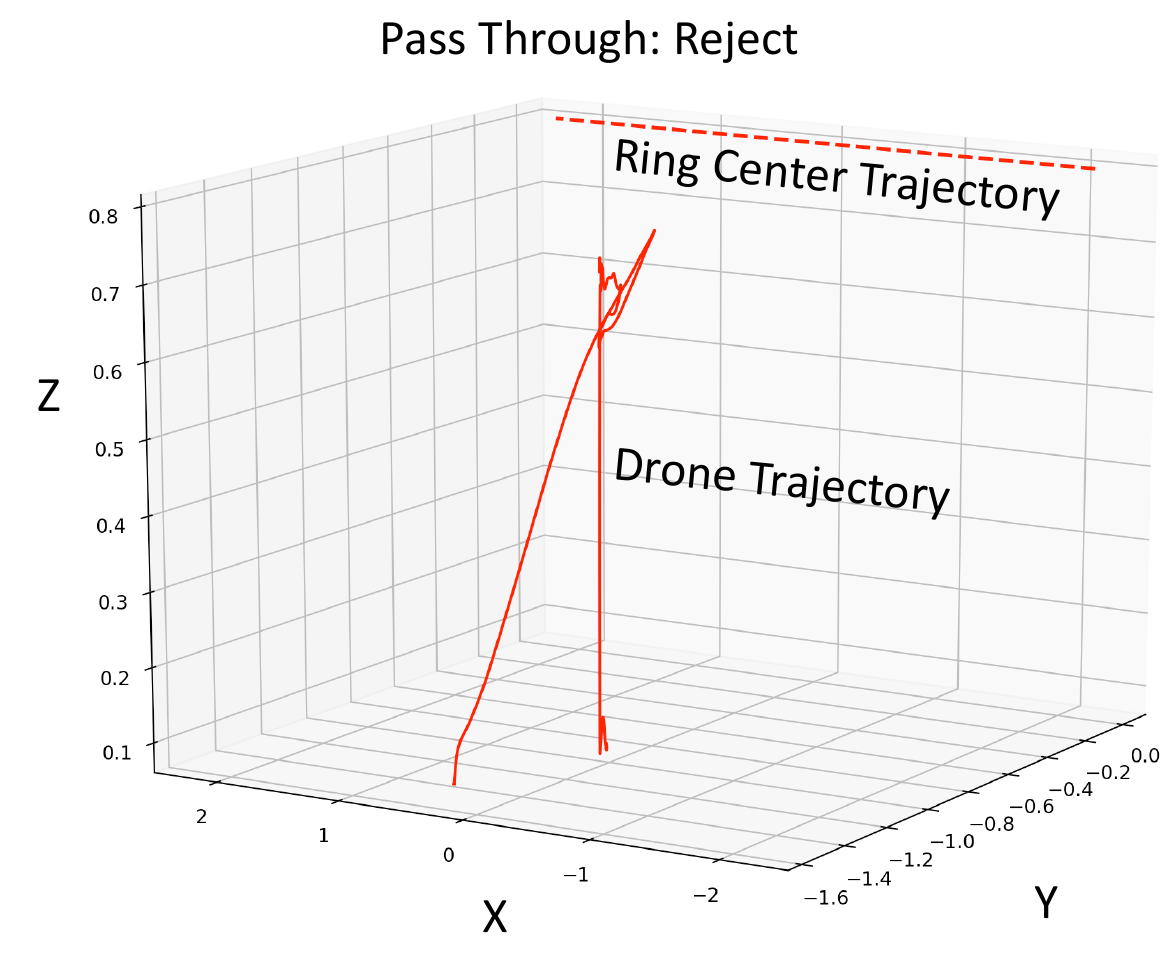} 
    \caption{Drone trajectory for the reject case showing immediate landing after flight deemed infeasible.}
    \label{fig:reject_case}
\end{figure}

\begin{figure*}[!t]
    \centering
    \begin{subfigure}{0.32\textwidth}
        \includegraphics[width=\textwidth]{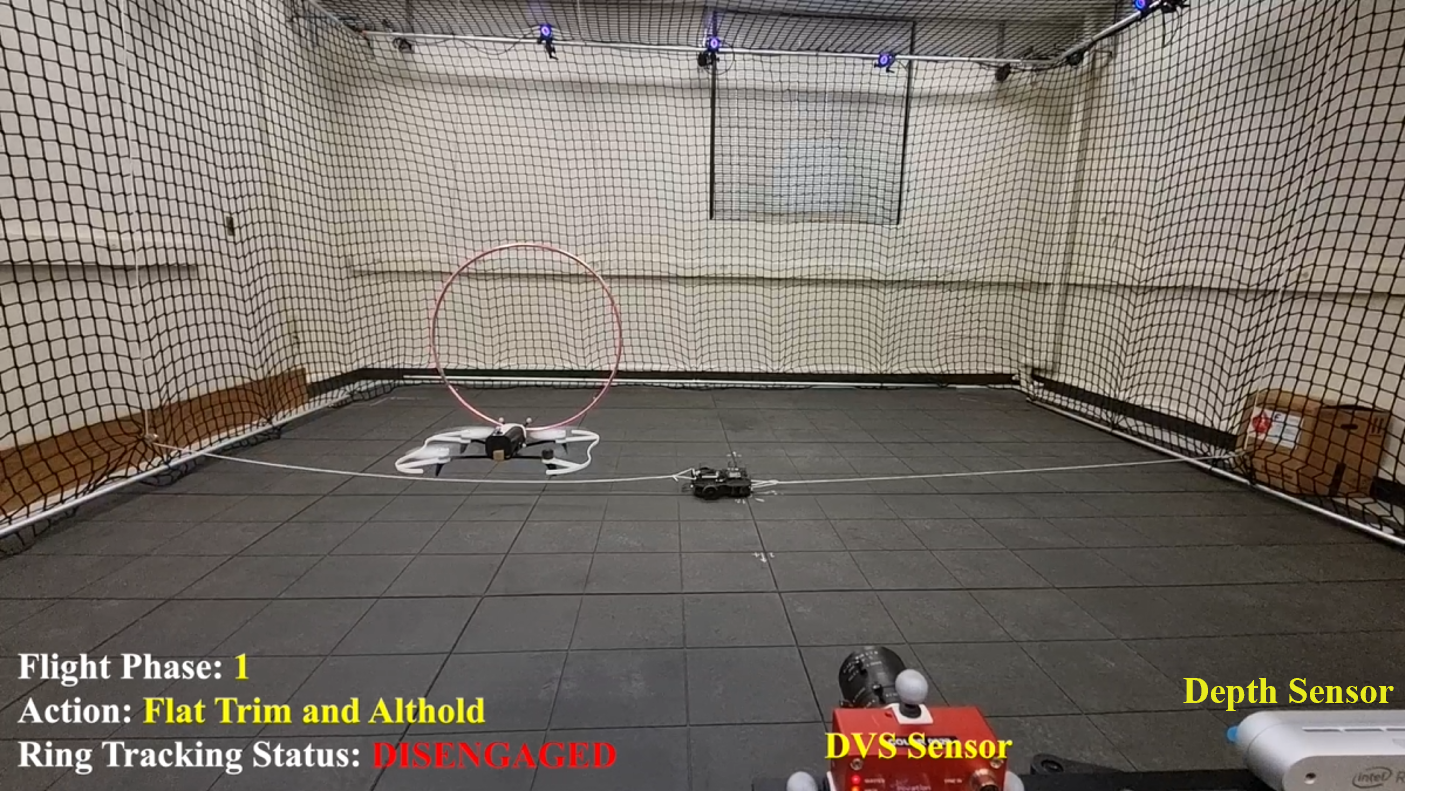}
        \caption{Phase 1: Flat Trim and Althold}
        \label{fig:st1}
    \end{subfigure}
    \hfill
    \begin{subfigure}{0.32\textwidth}
        \includegraphics[width=\textwidth]{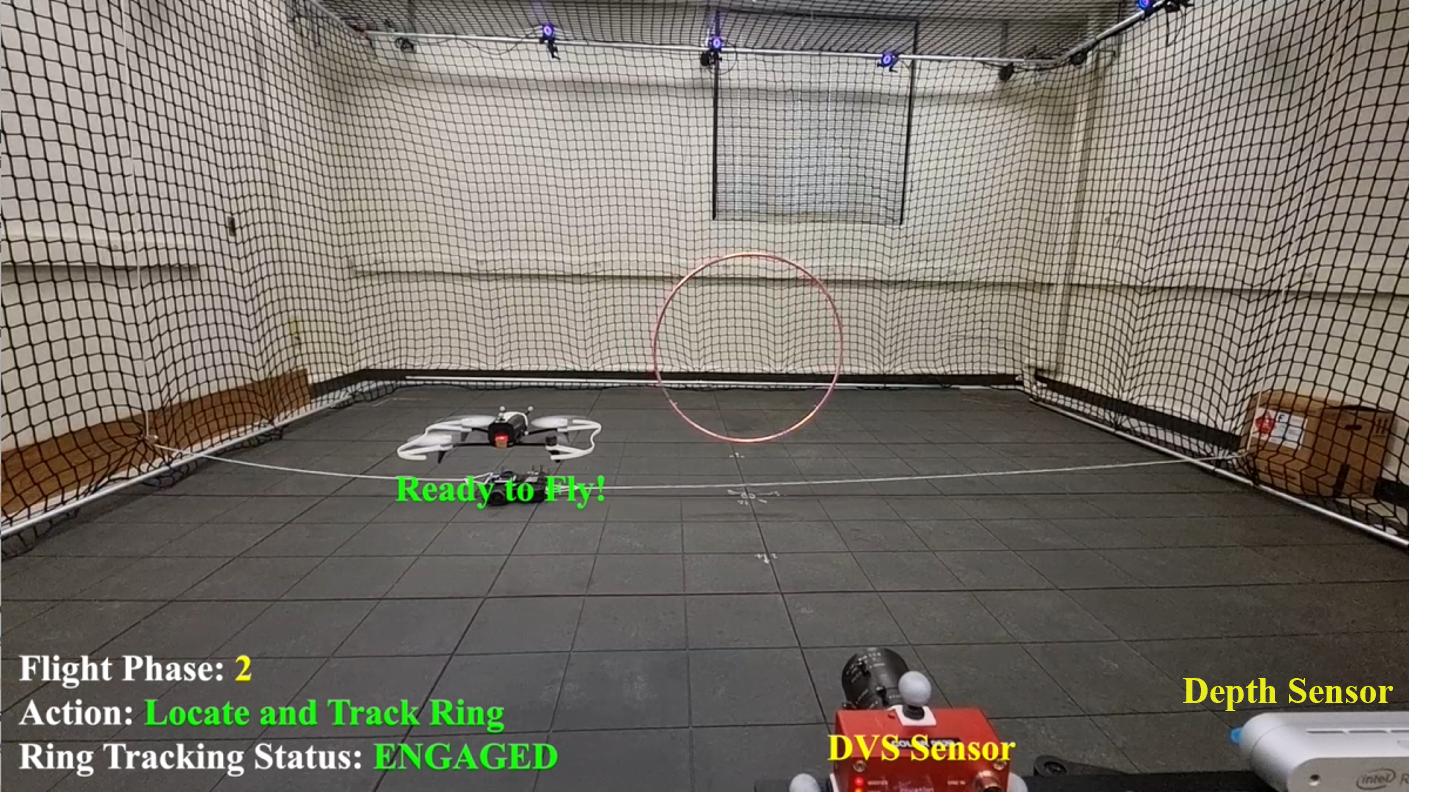}
        \caption{Phase 2: Locate and Track Ring}
        \label{fig:st2}
    \end{subfigure}
    \hfill
    \begin{subfigure}{0.32\textwidth}
        \includegraphics[width=\textwidth]{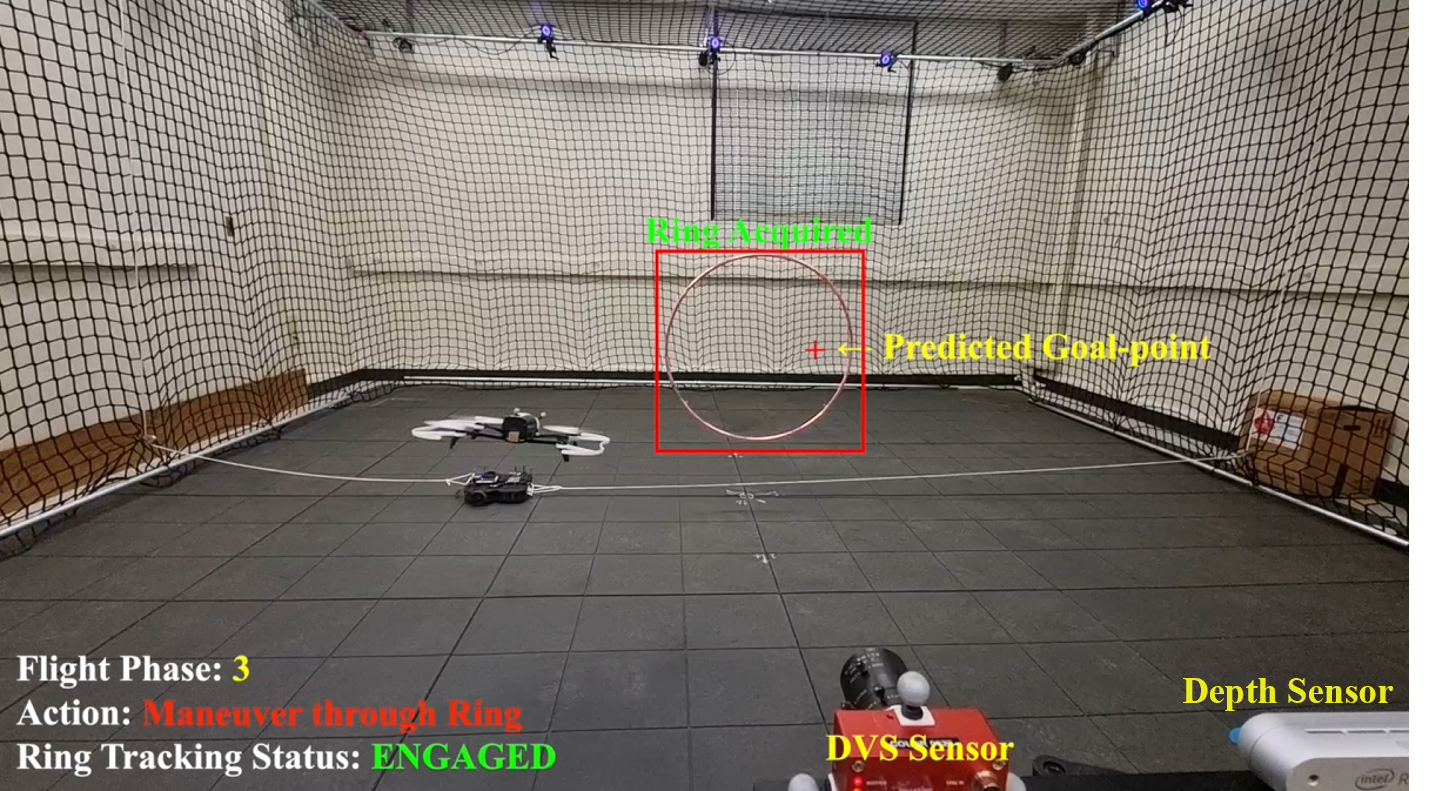}
        \caption{Phase 3: Maneuver through Ring}
        \label{fig:st3}
    \end{subfigure}
    \hspace{177.80093mm}
    \begin{subfigure}[t]{0.32\textwidth}
        \includegraphics[width=\textwidth]{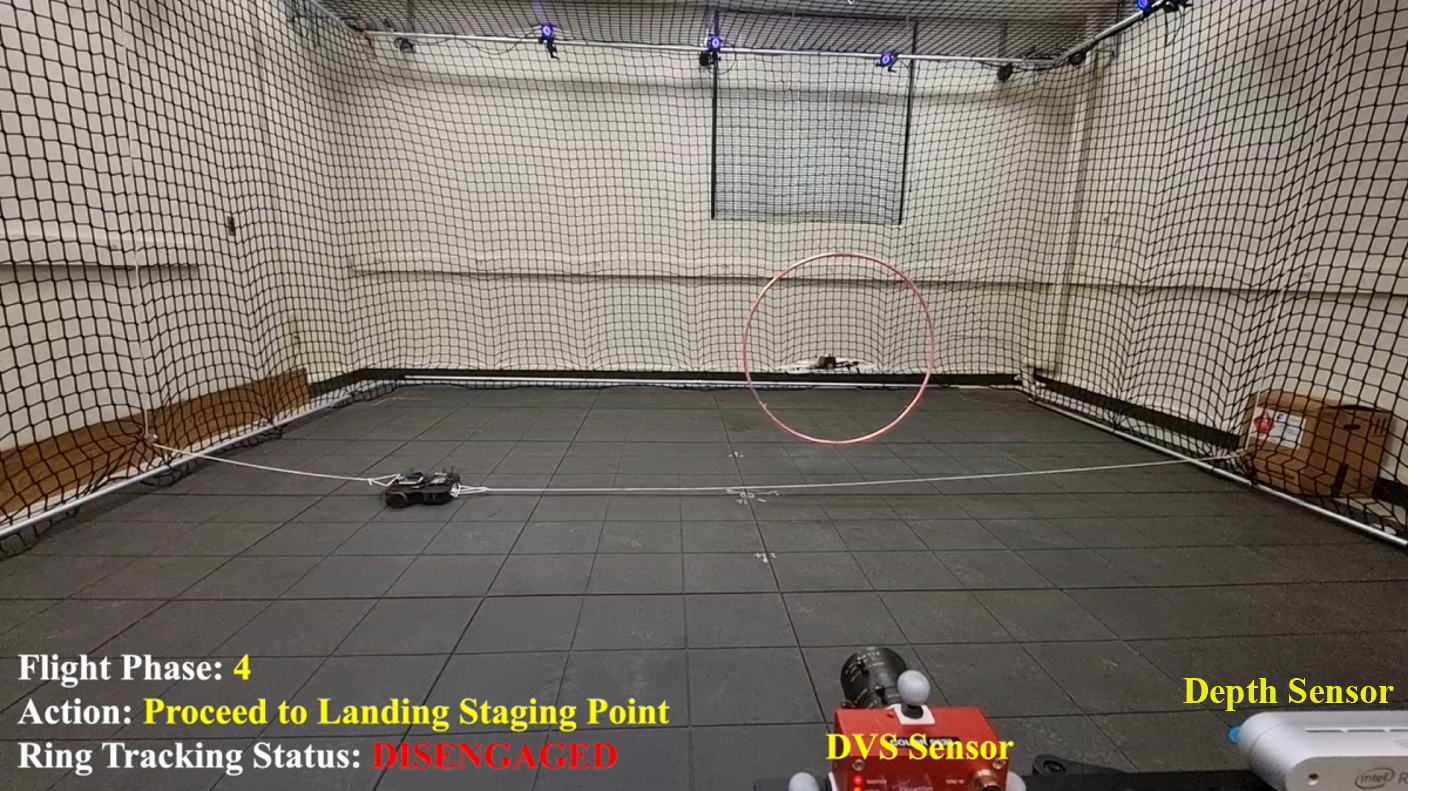}
        \caption{Phase 4: Prepare to Land}
        \label{fig:st4}
    \end{subfigure}
    \hspace{1mm}
    \begin{subfigure}[t]{0.32\textwidth}
        \includegraphics[width=\textwidth]{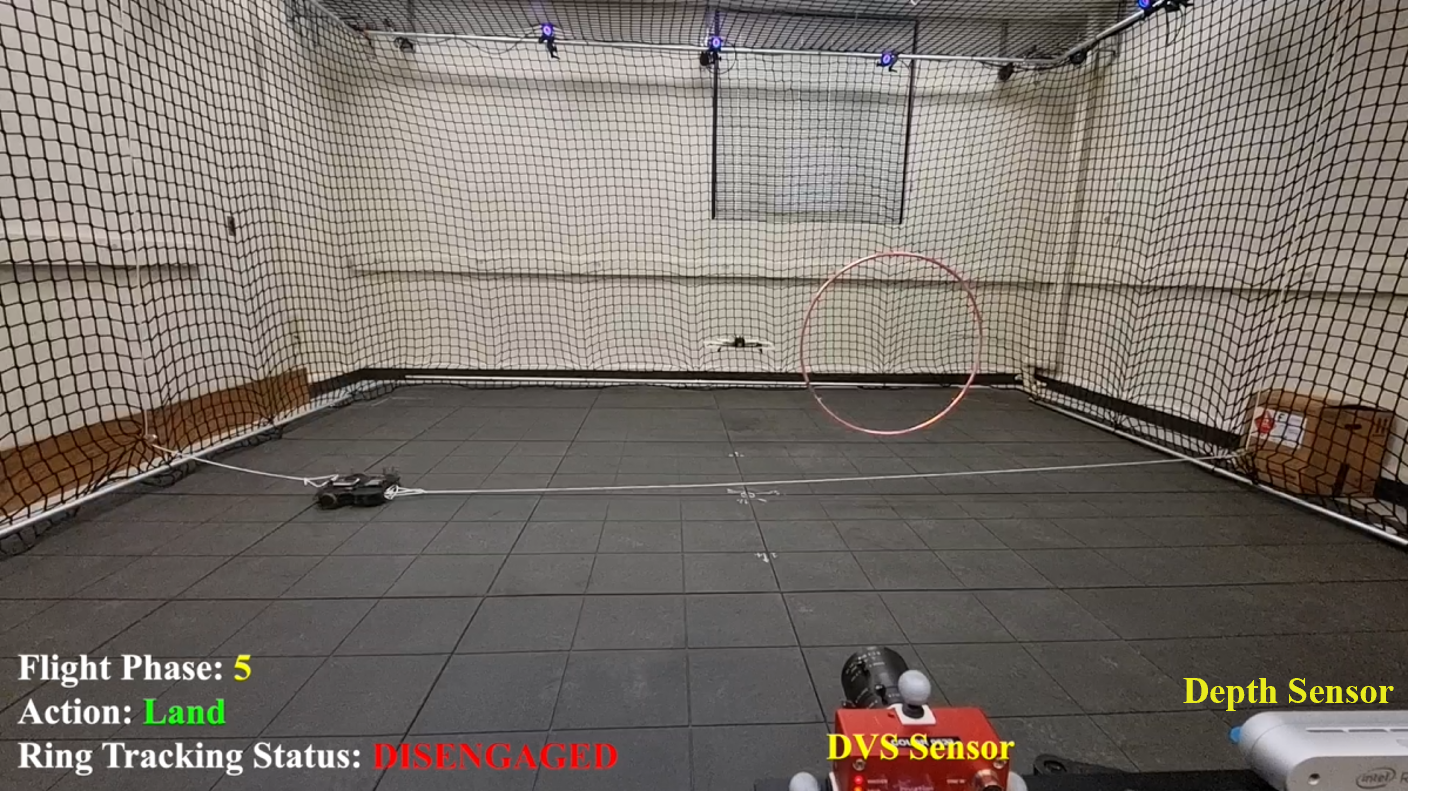}
        \caption{Phase 5: Land}
        \label{fig:st5}
    \end{subfigure}
    \caption{Phases of Neuro-LIFT Control Execution for "Fly through Center of Ring"}
    \label{fig:maneuver_phases1}
\end{figure*}

\subsection{\textbf{LLM-based Neuromorphic Drone Flight}}

This section presents a comprehensive performance evaluation of the proposed Neuro-LIFT framework, which integrates neuromorphic sensors and large language models (LLMs) to enhance autonomous drone navigation. The framework demonstrates significant advancements in terms of robustness, energy efficiency, and adaptability, particularly in dynamically changing environments where traditional systems might falter.

Table \ref{tab:llm_instructions} outlines the high-level commands translated by the Large Language Model (LLM) into actionable tasks for the drone. These instructions are critical for enabling seamless human-drone interaction, effectively reducing the complexity of command interpretation and execution in real-time operational scenarios. The LLM's advanced natural language processing capabilities allow it to understand and execute commands that require nuanced understanding of the physical environment, which is instrumental in achieving precise and safe navigation outcomes.

The instructions listed in the table encompass a range of navigational commands, from basic maneuvers like "Fly Left of Ring" to more complex tasks such as "Fly Through Ring at Different Orientations." Each command is paired with potential responses from the drone, reflecting either a successful execution or a need to reject the maneuver based on safety algorithms and current flight dynamics. 

Figure \ref{fig:drone_trajectories} provides a visual representation of the drone's flight paths under varying navigational commands. The 3D trajectories illustrate how the drone adapts its flight path in real time to the commands processed by the LLM. These paths show the drone's ability to dynamically adjust its trajectory to avoid obstacles and navigate through confined spaces, which is crucial in environments where static flight paths would be inadequate. The figure highlights scenarios such as 'Go Around' and 'Pass Through,' where the drone demonstrates agility and precision in avoiding or navigating through a moving ring, respectively.

Figure \ref{fig:reject_case} illustrates the reject response mechanism within the Neuro-LIFT system. This feature is vital for preventing the execution of unsafe or unfeasible maneuvers. The illustrated reject case, where the drone deviates from its intended path and fails to pass through the ring, highlights the system's built-in safety measures that trigger a flight path correction or system halt to avoid potential accidents.

\subsection{\textbf{Neuro-LIFT Control Execution}}

Figure \ref{fig:maneuver_phases1} illustrates a typical flight maneuver performed by the drone during a successful navigation task:

\begin{enumerate}
    \item \textbf{Phase 1: Flat Trim and Alt-Hold} \\
    The drone stabilizes in a hovering state, undergoing an initial calibration to ensure reliable IMU readings. This step is essential for stable flight dynamics and accurate navigation.

    \item \textbf{Phase 2: Locate and Track Ring} \\
    The drone's neuromorphic sensing module, using inputs from the DVS sensor, identifies and tracks the ring. This phase highlights the advantages of event-based vision systems in dynamic environments, showcasing reduced latency and energy usage.

    \item \textbf{Phase 3: Maneuver Through Ring} \\
    The drone follows a computed trajectory to navigate through the ring, adjusting its path dynamically with a low-level PID controller. This tests the integration of neuromorphic computing with traditional control systems in real-time obstacle navigation.

    \item \textbf{Phase 4: Prepare to Land} \\
    After navigating through the ring, the drone begins to reduce altitude and adjust speed, preparing for landing. This phase requires precise control to transition smoothly from navigation to landing.

    \item \textbf{Phase 5: Land} \\
    The drone performs a controlled landing, marking the end of the maneuver. 
\end{enumerate}
\vspace{1mm}

%% file: sections/lit_review.tex
Human-robot interaction has been one of the main research problems of the past decade, but until very recently, was restricted to conventional NLP techniques. In the last few years, the rise of LLMs has spurred research in the area significantly. Zhang, et al. \cite{10546317} experimented with transformers in robot decision-making, showing that LLM-based systems achieve higher accuracy and faster response times in complex HRI scenarios than NLP-based approaches. Cui, et al. \cite{cui2024survey} conducted a survey demonstrating how multimodal LLMs outperform traditional NLP models in combining language with sensory inputs like images and LiDAR for autonomous driving and HRI tasks. \cite{sanyal2024asma} leverage vision-language models (VLMs) for scene-aware UAV navigation focusing on vision-language grounding, autonomous decision-making, and safety control. 

Bhat, et al. \cite{bhat2024hifi} explored how LLMs use few-shot learning to understand rare or ambiguous commands in HRI, a task that traditional NLP often fails to handle due to limited training data. Kim, et al. \cite{kim2024lingo} presented an LLM-based framework capable of converting natural language commands into executable robotic tasks, offering seamless human-robot communication compared to rigid NLP pipelines. However, this work is restricted to a quadruped robot (Boston Dynamics Spot) where power efficiency and agility are comparatively unimportant metrics. Additionally, these works focus on creating end-to-end pipelines, which significantly degrades the explainability of the networks.

In recent years, physics-based AI has proven powerful in embedding physical system knowledge into neural network learning processes, enhancing model robustness and adaptability. RAMP-Net \cite{sanyal2023ramp} uses physics-informed neural networks to boost control accuracy and adaptability in quadrotors. Building on these innovations, the EV-Planner framework \cite{sanyal2024ev} merges neuromorphic vision sensors and physics-guided neural networks, addressing traditional sensing delays and optimizing actuator energy use with hardware demonstrations \cite{joshi2024}. This enhances endurance and efficiency in resource-constrained environments.

%% file: sections/conclusion.tex
We proposed the Neuro-LIFT framework, a neuromorphic, LLM-based interactive system designed for autonomous drone navigation. The framework has been rigorously tested and has demonstrated robustness, adaptability, and efficiency in real-time operational environments.
Our experiments demonstrated the framework's capability to handle complex autonomous maneuvers with precision and reliability. The integration of neuromorphic vision with spiking neural networks played a crucial role in enhancing the drone’s responsiveness to dynamic changes in the environment, significantly reducing latency and power consumption while maintaining high accuracy in object tracking and trajectory planning.
Trajectory analysis demonstrated the system’s proficiency in adapting to different scenarios, successfully avoiding obstacles or navigating through them with precision. The ability of the Neuro-LIFT to navigate through a dynamically moving ring in multiple test setups highlighted its practical applicability in scenarios that demand high agility and precise real-time decision-making.
Importantly, the reject cases served a critical safety function by preventing the execution of maneuvers when flight conditions were not feasible. This aspect of the system enhances safety and reliability, ensuring that the drone only engages in flight paths that are secure and manageable under current conditions.
Neuro-LIFT framework merges advanced LLM technology with neuromorphic computing to achieve a new level of interaction and autonomy in drone navigation. This  (hopefully) establishes a foundation for future innovations in integrating large language models with real-time, physics-based neuromorphic systems.